\documentclass[letterpaper, 10 pt, conference]{IEEEtran}  
\usepackage{graphicx}
\usepackage{algorithm}
\usepackage{mathtools}

\usepackage{algorithm}
\usepackage{algorithmic}
\usepackage{mathrsfs}
\usepackage{amsfonts}
\usepackage{epsfig} 
\usepackage{mathptmx} 
\usepackage{amsmath} 
\usepackage{amssymb}  
\usepackage{multicol} 
\usepackage{upgreek}
\usepackage{bm} 
\usepackage{cite}
\usepackage{subeqnarray}
\usepackage{multirow}
\usepackage{epsfig} 
\usepackage{mathptmx} 
\usepackage{times} 
\usepackage{booktabs}
\usepackage{sansmath}

\begin{document}


\title{Bio-inspired Adaptive Latching System for \\ Towing and Guiding Power-less Floating Platforms with Autonomous Robotic Boats}

\author{\IEEEauthorblockN{Luis A. Mateos, Wei Wang, Fabio Duarte}
	\IEEEauthorblockA{MIT  \\
	}
}




\maketitle
\thispagestyle{empty}
\pagestyle{empty}


\begin{abstract}

Autonomous robotic boats are expected to perform several tasks: 1) navigate autonomously in water environments, such as the canals of Amsterdam; 2) perform individual task, such as water monitoring, transporting goods and people; 3) latch together to create floating infrastructure, such as bridges and markets. 

In this paper we present a novel bio-inspired robotic system for  latching, towing and guiding a floating passive-power-less  platform. 
The challenge is to design an adaptive latching mechanism, able to create a secure connection between the entities, easy to attach/detach, even if the boats are affected by water disturbances. 
But most important, the adaptive latching must be able to restricting the DoF (degrees of freedom) of the latched "dummy" platform. Since, the robotic boat may drive it in narrow water canals and must prevent it from drifting and hitting the wall. 




This novel adaptive latching mechanism is based on the ball and socket joint that allows rotation and free movements in two planes at the same time. It consists of two parts: the male part that includes a bearing stud (ball) integrated on the floating bin "dummy" and the female part located on the autonomous robotic boat. Which integrates an adaptive framed funnel to guide the male ball into an actuated receptor that traps the ball, creating the ball-socket joint between the boats. In this sense, the adaptive latching mechanism mimics squid's tentacles that can adjust the forces applied to a holding object restricting its degrees of freedom.

Experimental results are presented from our swarm robotic boats integrating the adaptive latching system and performing the towing and guiding use cases.


\end{abstract}

\section{Introduction}

Similar to the emerging self-driving cars, autonomous robotics boats are capable to sense their surroundings and navigate without human input \cite{7350466} \cite{7830353} \cite{8441797}. 
Autonomous robotic boats can be used for transporting goods and people on canals or rivers around urban cities, as well as monitor the coastal city's waters environment \cite{5584552} \cite{8447040} \cite{8460632}.

\begin{figure}[t]
	\centering
	\includegraphics[width=0.47\textwidth]{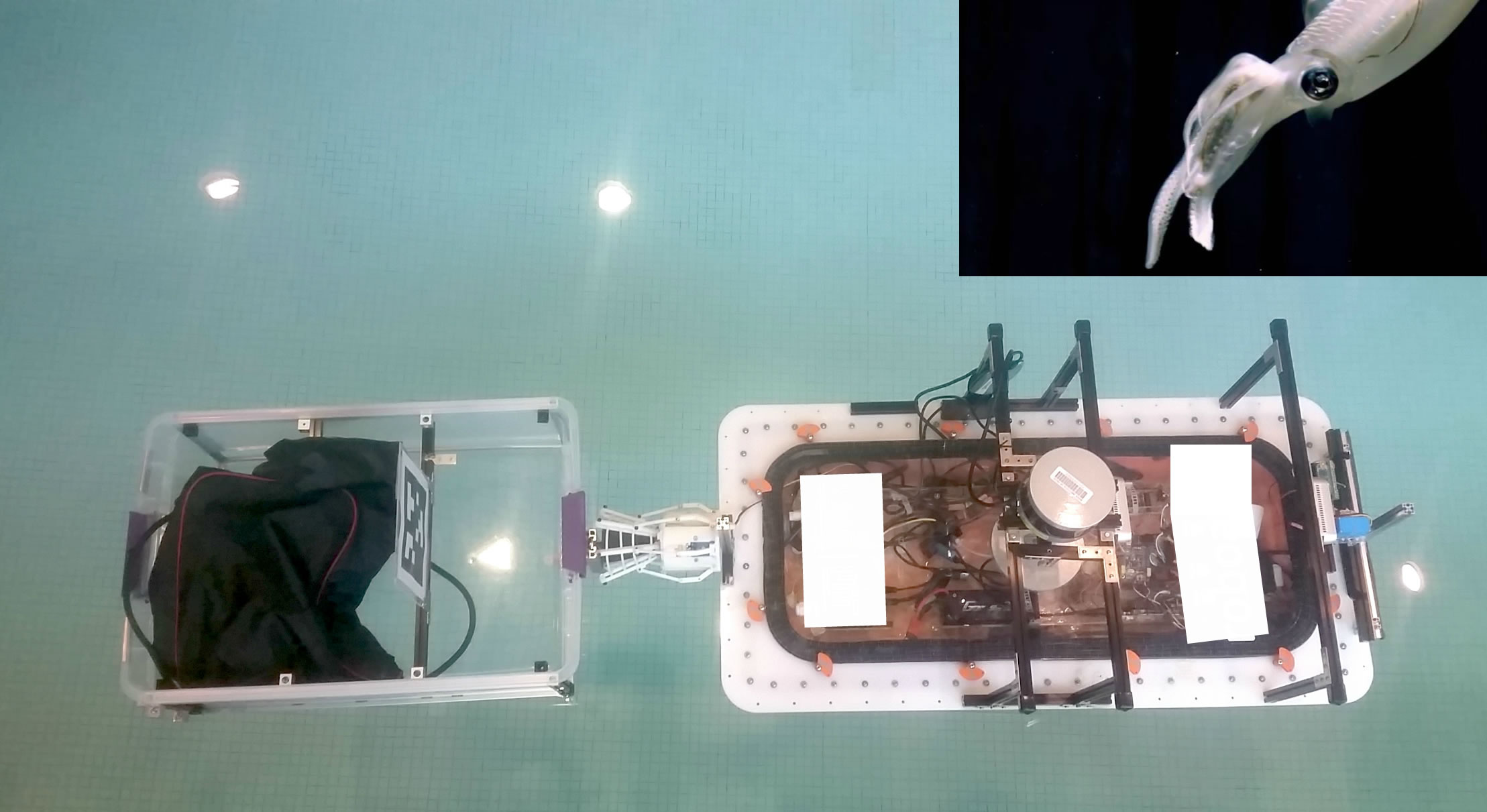}
	\caption{Autonomous robotic boat towing and restricting the DoF of a latched "dummy" container. The robotic boat integrates an adaptive funnel to guide the passive container, mimicking squid's tentacles when holding an object restricting its DoF (top-right).}
	\label{figintro}
\end{figure}

One of the biggest challenges of boats with autonomous mobility is to dynamically connect and disconnect them securely and efficiently to create floating infrastructure, such as bridges, markets or concert stages. Additionally, these autonomous robotic boats are not only required to latch to a docking station or to other robotic boat, as they must be able to latch a "dummy" floating platforms to tow it, see Figure \ref{figintro}. Once the robot latches the passive entity, it must be able to limit the degrees of freedom (DoF) of the power-less  boat when towing it in narrow canals, so the passive-power-less entity navigates the canal without drifting and hitting on the walls. After the latched entities pass the narrow canals, the DoF can be increased for a better maneuverability in the open water or when making a turn.



The presented latching system is based from autonomous underwater vehicles (AUV) docking stations, since these have an extensive research and practical developments  \cite{book1}\cite{umobiledock}\cite{electromag}\cite{remus1}\cite{21}\cite{visual}. On the other hand, autonomous surface vehicles (ASV) research has been relatively limited in the topic of latching systems for towing and guiding passive power-less floating platforms \cite{948815}\cite{44444}\cite{6907011}\cite{8793525}.

\textbf{Latching ASV: }
The process of latching includes several steps, the first step is the $guiding$, meaning how the entities are guided towards each other \cite{book1}\cite{umobiledock}\cite{fuzzy}.  
The second step for latching is the $docking$, which specifically refers to join one or more entities between them 
 \cite{optimal}\cite{oterm}\cite{servoing}\cite{fuzzy}.  
The third step is the $garaging$ in which the connection between the entities is locked. Common approaches are push \& lock mechanisms and motorized-screws, used in reconfigurable robotics \cite{6766548}.  

For the presented research we integrated the camera-tag framework for $guiding$, due to its high accuracy. The $docking$ stage integrates an adaptive framed funnel, which is a tubular garage commonly shaped as a cone, that helps minimizing the level of precision required to dock by increasing the target size \cite{oterm}\cite{servoing}\cite{fuzzy}. 
The $garaging$ stage creates a ball-socket joint between the entities with a motorized-socket. 

 Latching mechanisms for robotic boats are able to latch multiple boats with a hook-wire mechanism \cite{6907011} \cite{7091046}. However, these latching systems require that all boats are powered, since actuators are required on each boat. Making it impossible to latch and guide a completely power-less boat. 



\begin{figure}[t]
	\centering
	\includegraphics[width=0.49\textwidth]{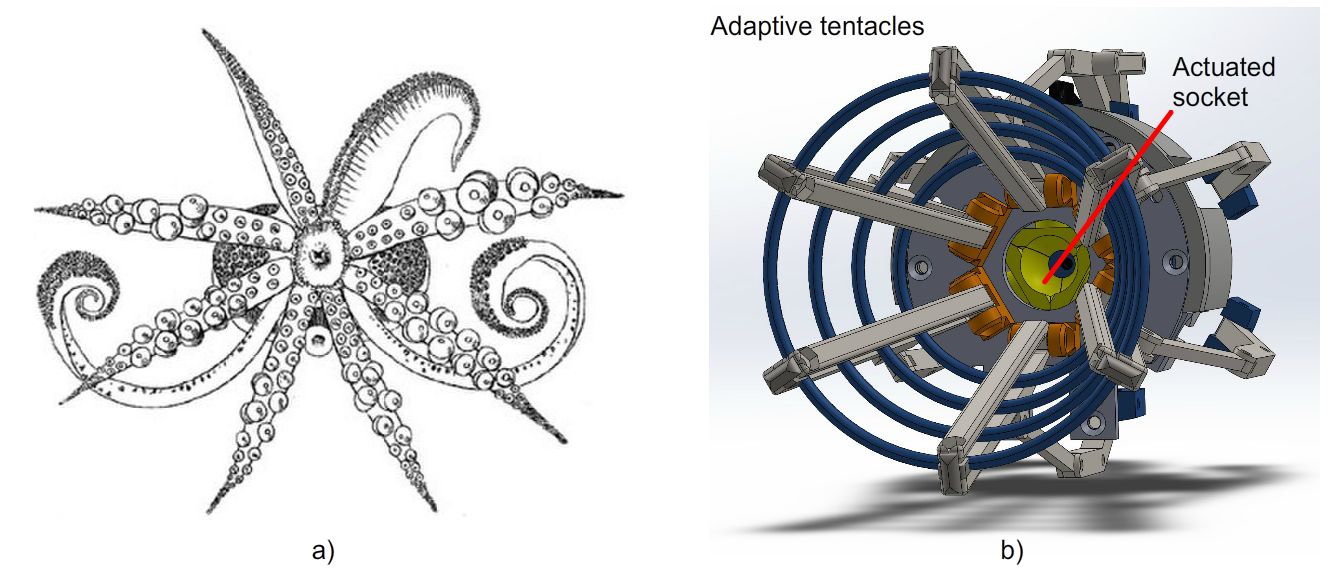}
	\caption{a) Oral view of the bobtail squid Semirossia tenera. b) Adaptive latching structure with dynamic framed funnel and actuated receptor.}
	\label{figbio}
\end{figure}

In this paper we present a novel adaptive latching mechanism that enables robotic boats to dynamically create floating infrastructure securely and efficiently. 
The proposed latching mechanism compensates and overcomes the misalignment from water disturbances with an adaptive funnel. Further, this dynamic structure mimics squid's tentacles and its mount when holding and "eating" an object, see Figure \ref{figbio}.  
Furthermore, this dynamic mechanism acts as an adaptive damping system for adjusting the stiffness of the latched entity, in order to achieve different functional modes, when towing and when creating floating structure. 


This paper is structured as follows:  
Section \ref{sec:design} describes the autonomous robotic boats  and the design of the adaptive funnel. 
Section \ref{sec:model} presents the bio-inspired model. 
Section \ref{sec:simu} describes the framework and functional modes of the latching system, when restricting 1DoF and 2DoF for towing and guiding a "dummy" boat. 
Section \ref{sec:experiments} shows the experimental results. 
Conclusions are proposed in Section \ref{sec:conclusion}. 


\section{Design}
\label{sec:design}

The robotic boat consists of a rectangular base (2:1 ratio) with four thrusters in the middle of its edges. In this relationship the robot is able to move forward, backward, sideways and able to rotate on its axis. The dimension of this robotic platform are $1000mm \times 500mm \times 150mm$.
The perception and location of the robot is performed by a 3D lidar VLP16 (16 lines) located at the top of the boat. Also, the robotic boats integrates a camera to detect tags in the docks or on other boats for identification and pose estimation. The system also integrates an Inertial Measurement Unit (IMU) for detecting the inclination and  velocities, plus a RTK GPS to improve the outdoor localization.



\begin{figure}[h]
	\centering
	\includegraphics[width=0.4\textwidth]{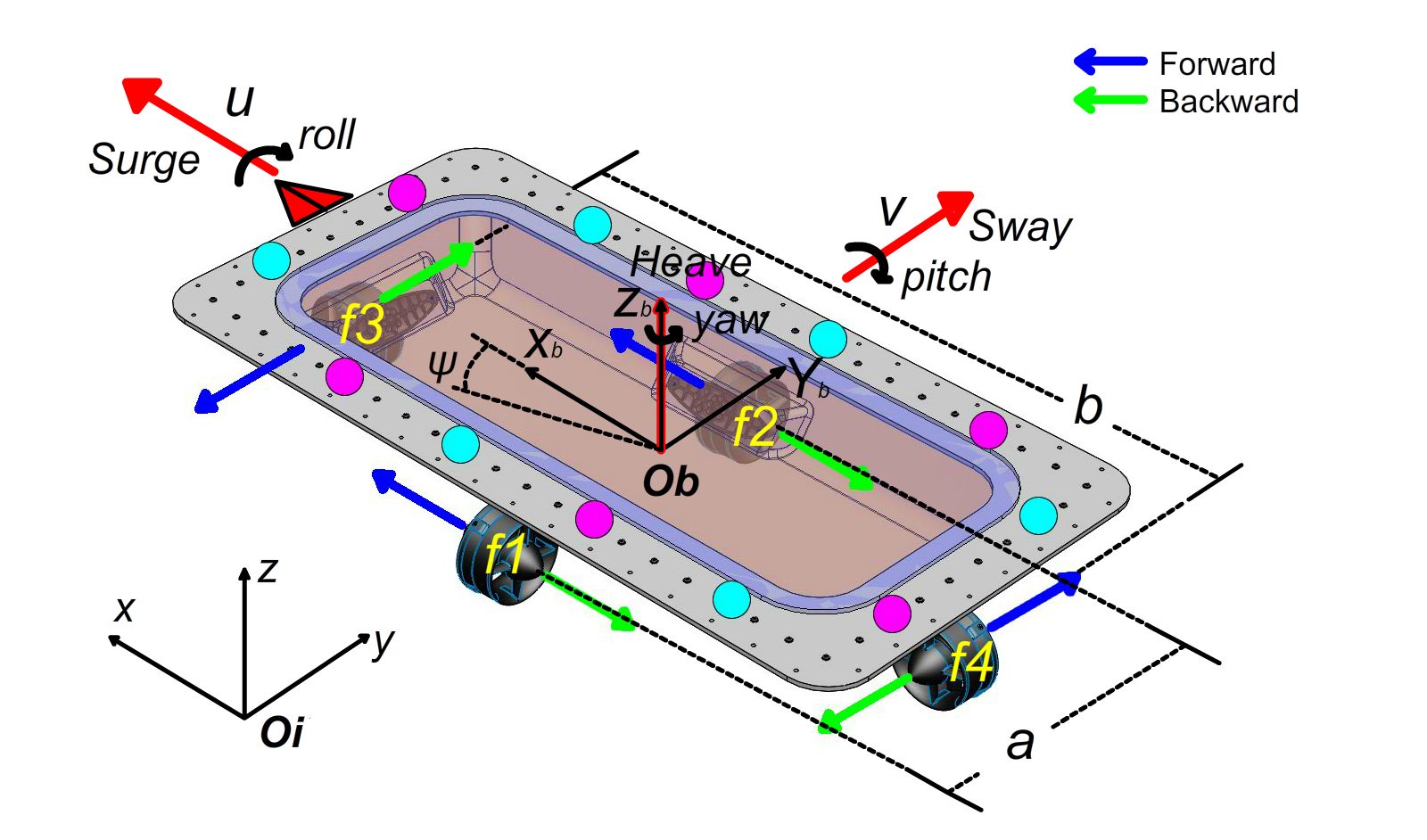}
	\caption{Robotic boat force vectors and coordinate frame. 
	}
	\label{figroboat2}
\end{figure}

The applied force and moment vector $\bm{\uptau}$ can be written as
\begin{eqnarray}\label{AppliedForceMaxtrix}
\bm{\uptau}
=\mathbf{B}\mathbf{u}
=
\left[
\begin{array}{cccc}
1                      &  1                                      &    0                       & 0\\
0                      &  0                                      &    1                       & 1\\
\dfrac{a}{2}&-\dfrac{a}{2}                 &     \dfrac{b}{2} &-\dfrac{b}{2}
\end{array}
\right]
\left(
\begin{array}{c}
f_1\\
f_2\\
f_3\\
f_4
\end{array}
\right)
\end{eqnarray}

where $\mathbf{B}$ is the control matrix describing the thruster configuration and $\mathbf{u}$ is the
control vector. $a$ is the distance between the transverse propellers and $b$ is the distance between the longitudinal propellers, $f_1$, $f_2$, $f_3$ and $f_4$ are the forces generated by the corresponding propeller, see Figure \ref{figroboat2}. Each propeller is fixed and can generate continuous forward and backward forces.

\subsection{Adaptive latching system design}
\begin{figure}[t]
	\centering
	\includegraphics[width=0.45\textwidth]{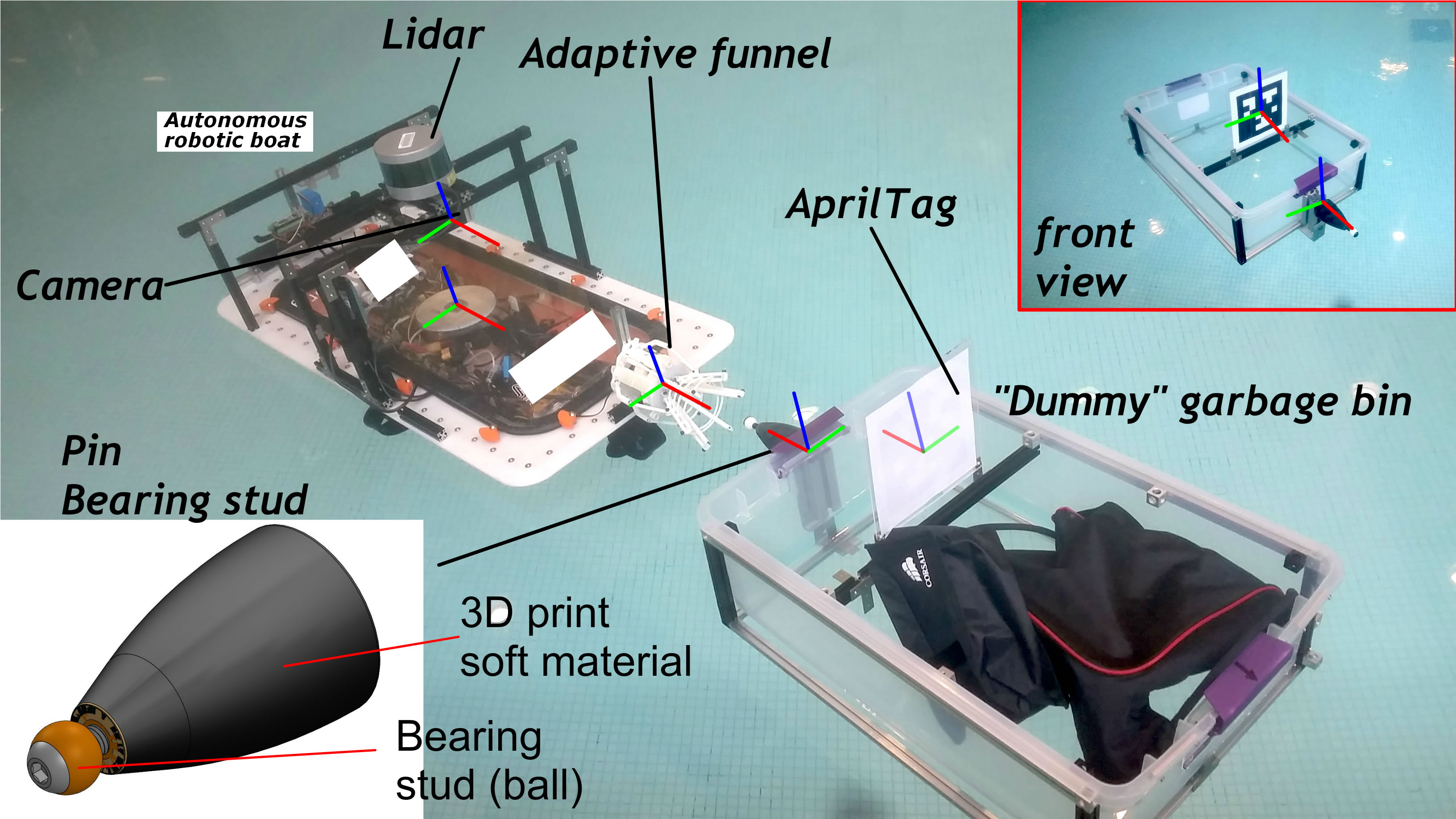}
	\caption{Autonomous robotic boat and power-less "dummy" boat with no sensor, no thruster, no actuators, only with a tag and a towing pin. 
	}
	\label{figroboat3}
\end{figure}

In order to latch on the water, our design integrates two gender entities: a pin with bearing stud (male), an adaptive framed funnel (female) and a guiding system to perform the latching, see Figure \ref{figroboat3}. The adaptive latching system for robotic boats is a bio-inspired concept from a squid with two main ideas: i) the adaptive framed-funnel mimics the tentacles of the squid for holding an object tightly or loosely and to guide it to its mouth. ii) The actuated receptor in the adaptive funnel is acting as the squid's mouth.

\subsection{Adaptive latching system elements}

The adaptive latching system consists of three elements:

\subsubsection{Pin with  bearing stud  (male)}

The male component of the system consists of a pin with a bearing stud on the front and a 3D printed soft material covering the pin for damping purposes, see Figure \ref{figroboat3}. The compression set of this soft-material rubber is $4-5\%$ and its polymerized density is $1.12-1.13 g/cm3$ see Table \ref{table3d}. 

The passive male part can be integrated in all entities of our framework: on the docking stations, on the "dummy" power-less boats and also on the robotic boats.


\subsubsection{Adaptive funnel (female)}
The female part consists of a framed funnel to guide the male ball into a receptor that traps the ball, creating the spherical joint between the parts. 

The adaptive funnel is integrated only on the robotic boats, since requires electric power for the servomotor. It consists of six arm with concentric rings to maintain the conical shape when opening or closing. The actuation of the funnel can be performed by one or more actuators pushing the arms back and forth. 

\begin{figure}[h]
	\centering
	\includegraphics[width=0.49\textwidth]{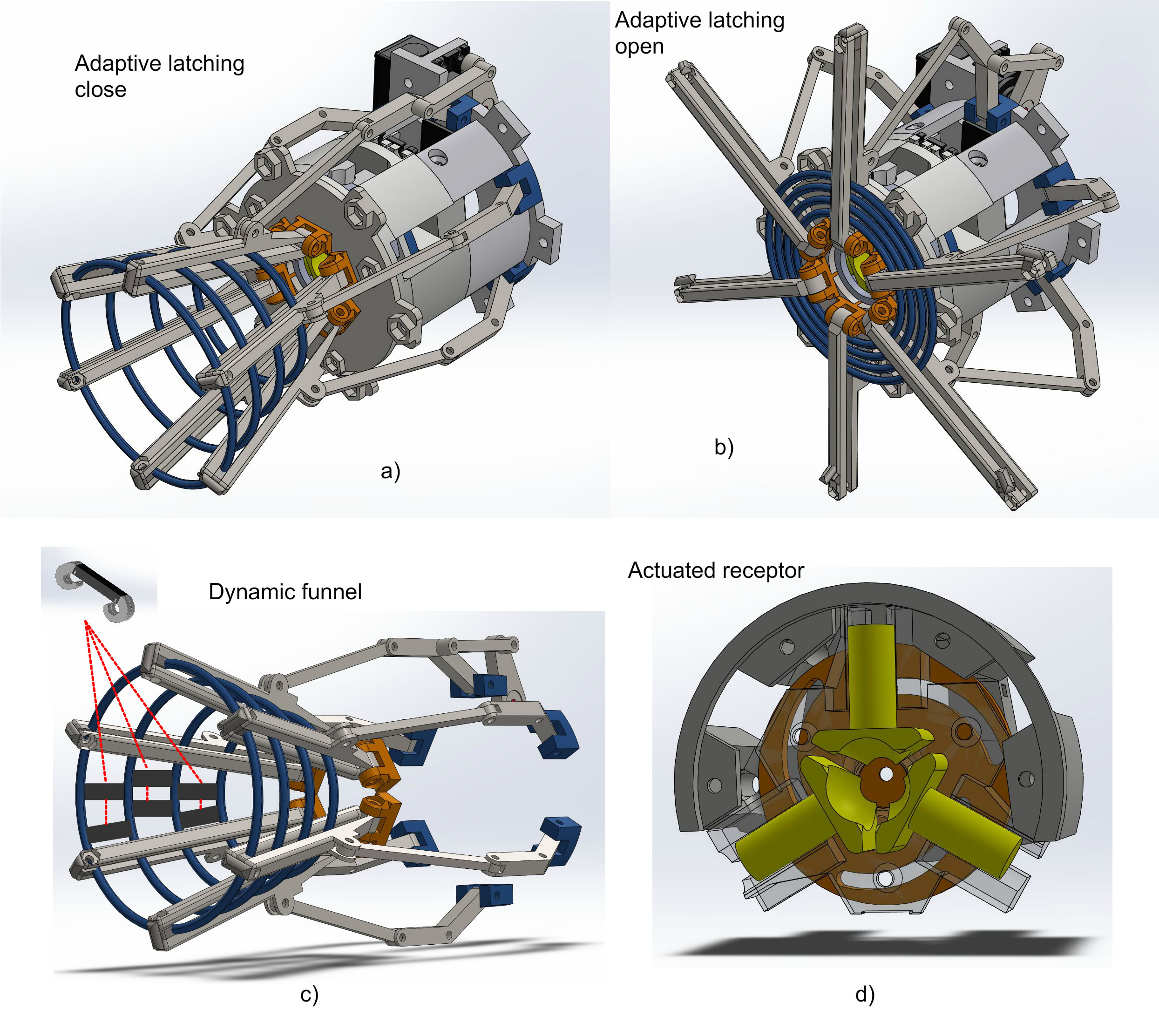}
	\caption{a) Adaptive latching with extended arms (close). 
		b) Adaptive latching with retracted arms (open). 
		c) Adaptive funnel structure with six movable arms and concentric rings. 3D printed elastic joints are set on the rings to follow the movements from the arms.
		d) Receptor mechanism with three arms that when closed forms the socket. 	}
	\label{figdfclosex2}
\end{figure}

The receptor is located inside the adaptive funnel, it consists of three arms that when closed form the socket. This receptor integrates a push and lock mechanism for detecting and trapping the bearing stud from the male element, see Figure  \ref{figdfclosex2}.

\textbf{Adaptive funnel 1 DoF. } 
	In order to actuate the adaptive funnel with 1 DoF, only one actuator is required. This can be mounted on the back of the structure and attached to one of the six connected arms. In this configuration, all the arms will follow homogeneously the movements from the actuated arm. 

\textbf{Adaptive funnel 2 DoF. } 
A couple of actuators are required to configure the adaptive funnel with 2 DoF. The actuators can be distributed opposite to each other and connected to the lead ring. In this way, the lead ring creates the shape of the funnel. If the linear actuators are completely extended then the funnel's diameter is minimal. On the other hand, if the linear actuator is compressed, then the funnel's diameter increases, opening the funnel, see Figure  \ref{figdfclosex2}. Moreover, the funnel can be set asymmetrical if the actuators are extended differently. 






\subsubsection{Passive markers}

In order to guide the autonomous robotic boat to latch the passive power-less floating vessel, we require a methodology that does not needs power on both entities. 
The robotic boat, which is powered can integrate a camera, while the "dummy" boat integrates a printed marker. 
For this reason, we adopted Apriltags for guiding the robot to latch the "dummy" boat. Since, these markers offer a good framework for identification and pose estimation \cite{7759617} \cite{5979561}, even in poor lighting conditions \cite{7271491}.

\section{Model}
\label{sec:model}

%
Conventional suspension consists of coil springs and dampers. Due to the fixed suspension settings, like the spring constant and damping ratio, the applicable range for vibration suppression is limited. Hence, in order to improve both vehicle ride and handling performance an active suspension system is required. 
Active suspension systems provide an extra force input in addition to possible existing passive systems by the incorporation of an actuator in parallel with the mechanical spring. As opposed to the passive control, active control can improve the performance over a wide range of frequencies.

 Similar to the automobiles active suspension, the presented adaptive structure acts as an active suspension system by damping forces perpendicular to the pin's central axis. There are a couple of damping factors in the adaptive frame: 1) fix damping factor from the pin's 3D printed soft material and 2) the actuator's force in each "tentacle" parallel to the soft cone shaped rubber, see Formula \ref{spring} and \ref{spring2}. 
 
\begin{figure}[h]
	\centering
	\includegraphics[width=0.49\textwidth]{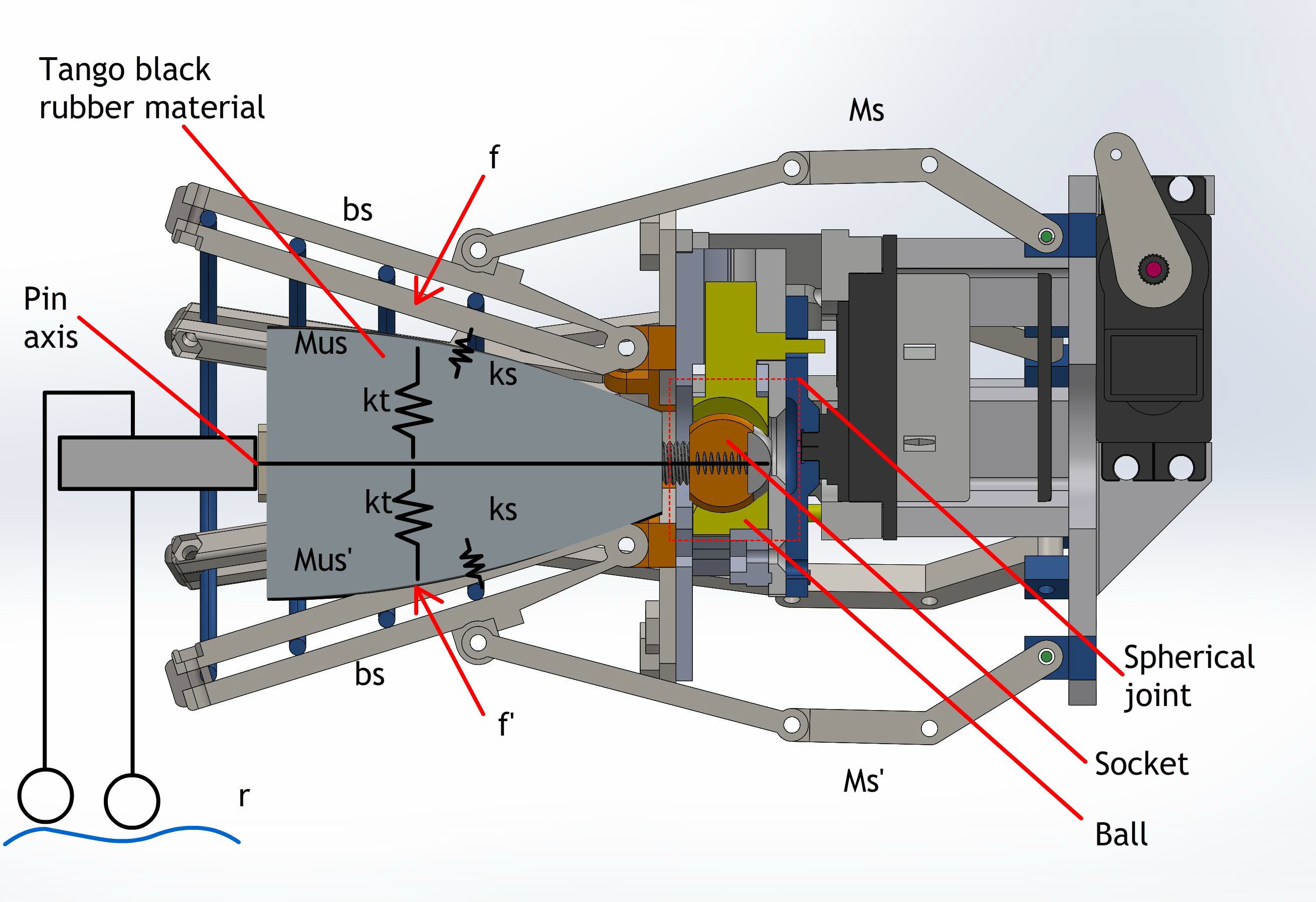}
	\caption{Adaptive funnel model. }
	\label{figpherical}
\end{figure}

 In our formulation and for simplification the 3D printed soft rubber can be in contact with only a couple of arms, see Figure \ref{figpherical}. The differential equations of motion for the two degree of freedom systems are:
 
 \begin{equation} \label{spring}
 \begin{split}
 M_s \ddot x_s = -k_s(x_s-x_{us}) - b_s(\dot x_s-f) \\
 M_{us} \ddot x_{us} = k_s(x_s-x_{us}) - b_s(f - \dot x_s) - k_t(x_{us}-r)
 \end{split}
 \end{equation}

\begin{equation} \label{spring2}
\begin{split}
M'_s \ddot x'_s = -k_s(x'_s-x'_{us}) - b_s(\dot x'_s-f') \\
M'_{us} \ddot x'_{us} = k_s(x'_s-x'_{us}) - b_s(f' - \dot x'_s) - k_t(x'_{us}-r)
\end{split}
\end{equation}

where,
\begin{itemize} 
	\item[] $M_s$ = Sprung mass (kg) 
	\item[] $x_s$ = Sprung mass displacement (m)
	\item[] $M_{us}$ = Unsprung mass (kg) 
	\item[] $x_{us}$ = Unsprung mass displacement (m)
	\item[] $k_s$ = Spring stiffness constant (N/m) 
	\item[] $b_s$ = Damping coefficient (Ns/m) 
	\item[] $k_t$ = Tyre stiffness constant (N/m) 
	\item[] $r$ = Road input (m) 
	\item[] $f$ = Actuator control force (N) 
	
	\item[] $M'_s$ = Sprung mass in opposite position (kg) 
    \item[] $x'_s$ = Sprung mass displacement in opposite position (m)
    \item[] $M'_{us}$ = Unsprung mass in opposite position (kg) 
    \item[] $x'_{us}$ = Unsprung mass displacement in opposite position (m)	
	\item[] $f'$ = Actuator control force in opposite position (N) 
\end{itemize}

\section{Vision-based controllers}
\label{sec:simu}


The visual controller consists of a camera and a tag for identification and pose estimation. 
We integrated squared fiducial marker systems, in specific the AprilTags for their good performance when detecting smaller markers, in high angle inclination and in different lighting levels \cite{7271491}.

Figure \ref{simbase} shows the AprilTags framework. If the camera is positioned in front of the tag, the lateral distance $d_y$ and angle $\psi$ are zero. While, the longitudinal distance between the camera and tag is $d_x$ in the $X-axis$. If the camera changes its position with the same orientation, then the lateral distance $d_y$ reflects the change in position on the $Y-axis$, $d_x$ is the same longitudinal distance and  $\psi=0$. If the camera in that position changes its orientation to face the tag, then the lateral distance $d_y=0$ and the angle reading becomes the true angle $\psi$ between the camera and tag.

\subsubsection{3D space simplification to 2D plane}
\begin{figure}[h]
	\centering
	\includegraphics[width=0.49\textwidth]{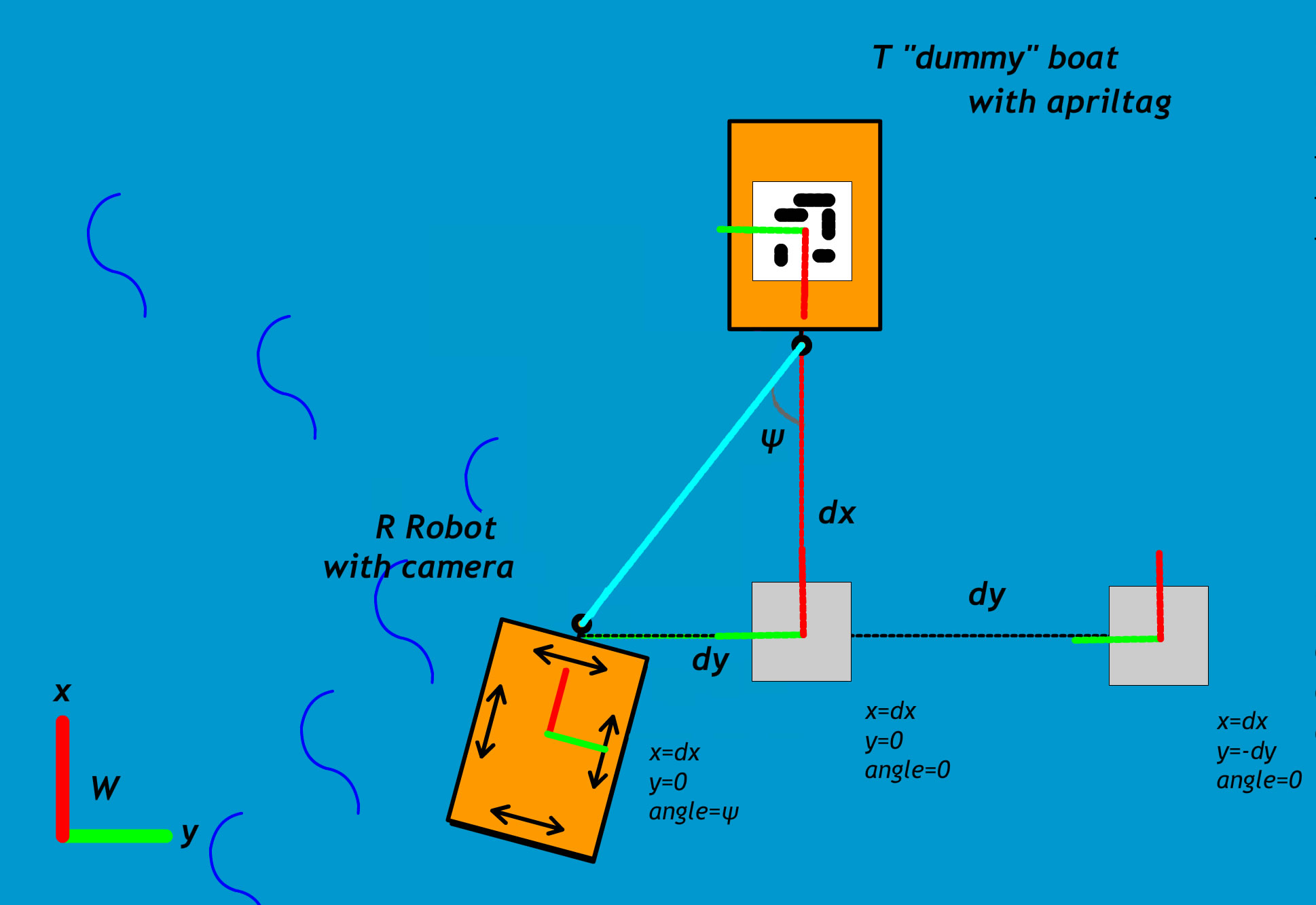}
	\caption{2D working space for latching, $d_y$ lateral distance, $d_x$ longitudinal distance and $\psi$ angle between the entities }
	\label{simbase}
\end{figure}

In our implementation we do not take into account the $heave$ vector as this is compensated by the funnel and in our assumption all elements are floating on the water. Thus, we simplify the 3D space to a 2D plane, see Figure \ref{simbase}.

\subsubsection{Hybrid controller}
The latching controller integrates four PD controllers:

\begin{itemize}
	\item Control I: Minimize lateral distance $dy_{R,T}=R_{dy}-T_{dy}$
	\item Control II: Min. longitudinal distance $dx_{R,T}=R_{dx}-T_{dx}$
	\item Control III: Minimize the angle $\psi$ between the entities 
	\item Control IV: Adaptive funnel acceptance degree 
\end{itemize}

The hybrid controller initially tries to set the robotic boat $R$ to have the same orientation with the target $T$ by minimizing  $dy_{R,T}$ and the angle $\psi$ between them. If the error is greater than the tolerances, then  the latching robot $R$ is set to maintain a distance of $d_X$ from the target to keep minimizing these errors, until the error from Control I and Control III are inside the funnel tolerances. In this case, the robot moves forward minimizing $dx_{R,T}$, while combining the lateral distance and angle strategies to latch to the target entity $T$. In this final stage, before reaching the target, Control IV is activated to start closing the funnel adaptively while closing its socket trapping the bearing-stud (ball) from the target entity, see Algorithm \ref{a1}.

The adaptive funnel may integrate one or more actuators to restrict the DoF of the latched entity. 

\begin{table}[b]
	\centering
	\begin{tabular}{|l|l|l|l|l|}
		\hline
		~                                     & Solid like-materials &  Rubber like-materials  \\ 
		~                                     & RGD-450 rigid &  FLX980 flexibe  \\ \hline
		Tensile strength              & 40-45 MPa & 0.8-1.5 MPa \\ \hline
		Elongation at break         & 20-35 \% & 170-220 \%\\ \hline
		Shore Hardness (A)         & Scale D 80-84 &  Scale A 26-28   \\ \hline
		Polymerized density        &   1.20-1.21 g/cm3  &  1.12-1.13 g/cm3  \\ \hline
	\end{tabular}
	\caption{}
	\label{table3d}
\end{table}

\subsubsection*{Adaptive latching with 1 DoF}

If the adaptive latching system integrates only one actuator, it commands all the "tentacles" to act in synchrony, acting as a global DoF. 
This is useful when the robotic boat is towing the "dummy" boat in a narrow canal, in which it must follow the robot as if it is connected rigidly to avoid any oscillation when navigating in this tight space, see Figure \ref{figdfclosexz}$a$.

In this scenario, the "tentacles" are closed grabbing strongly the pin in order to maintain the dummy's position exactly in the center, even if disturbances are affecting its position.

\subsubsection*{Adaptive latching with 2 DoF}

If the adaptive latching system integrates a couple of actuators for controlling the reconfigurable funnel (where each actuator is located  opposite to each other). Then, each of them can control half of the funnel's aperture angle. In other words, each actuator can control half of the framed funnel and can change the shape of the funnel asymmetrically.

\begin{figure}[h]
	\centering
	\includegraphics[width=0.48\textwidth]{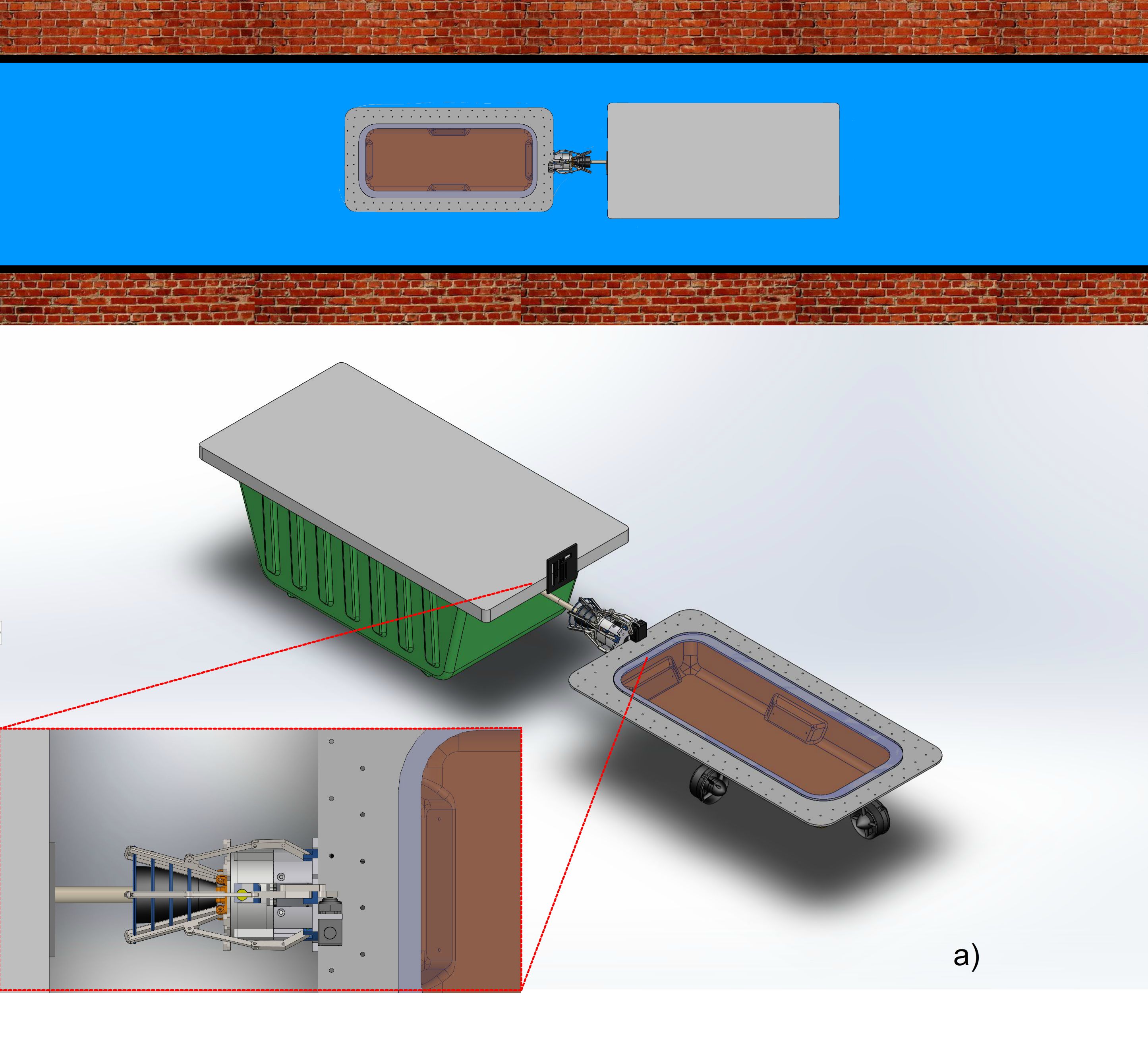}
	\includegraphics[width=0.48\textwidth]{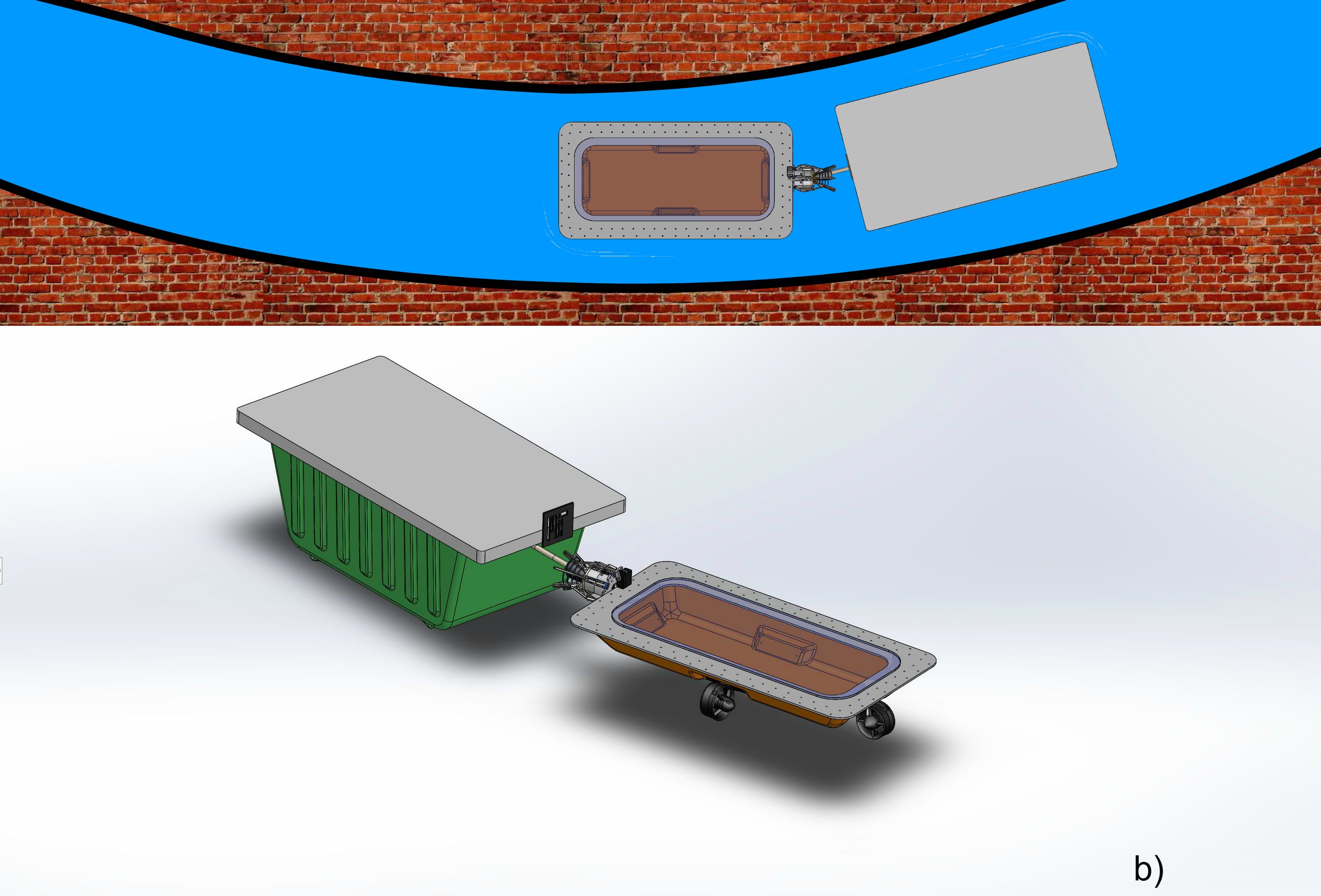}
	\caption{a) Closed adaptive funnel with 1 DoF mimicking a rigid connection for navigating in straight line in narrow canals. b) 2 DoF adaptive funnel able to direct the connected "dummy" boat to the  left or right, when navigating in canals with curvatures.}
	\label{figdfclosexz}
\end{figure}

Hence, the tentacles can guide a latched dummy boat to move to the left or to the right while navigating. This framework is useful when turning in a curvy narrow canal, see Figure \ref{figdfclosexz}$b$. 
In this configuration, the visual controller is not only used for detecting and latching, as it reads the true position and orientation of the towed entity and can adjust it accordingly.

\section{Experiments}
\label{sec:experiments}

This section contains the results of our bioinspired latching system tested indoors and outdoors. 
We present a couple of experiments: 1) the autonomous robotic boat latches a docked "dummy" on the swimming-pool and drives it restricting its DoF, we tested with 1 DoF and 2 DoF. 
2) the autonomous robotic boat latches and tows a "balancing" robotic boat that is floating on the river with 1 DoF. 

\begin{algorithm}[!h]
	\caption{Adaptive Latching Controller Algorithm }
	\label{a1}
	\begin{algorithmic} 
		
		\REQUIRE $dy_{2,1}, dx_{2,1}, \psi, flag$\_$missed$\_$target$,\\ $Adaptive$ $(funnel$ $state)$
		\ENSURE Camera targeting tag 
		\STATE min($dy_{2,1}$)
		\STATE  min($\psi$)
		
		\IF {$flag$\_$missed$\_$target == 0$ } 
		\IF {$dx_{2,1} > 0mm$ } 
		
		\IF {$dy_{2,1} < 10mm$ \OR $\psi < 2^\circ$}
		\STATE min($dx_{2,1}$)  
		\STATE $adaptive_{close}$ $(proportional$ $to$ $d_x$ AND $d_y)$
		\ELSE
		\STATE Move back $1m$ 
		\STATE  minimize $dy_{2,1}$ and $\psi$	
		\STATE min($dx_{2,1}$ - $1m$)  
		\STATE $adaptive_{open}$ $(proportional$ $to$ $d_x$ AND $d_y)$
		\ENDIF
		\ELSE 
		\STATE $flag$\_$missed$\_$target = 1$
		\ENDIF  
		\ENDIF
		
		\IF {$flag$\_$missed$\_$target == 1$ }
		\STATE Go to initial position and retry to latch
		\STATE min($dx_{2,1}$ - $1m$) 
		\IF {$dx_{2,1} > 1m$ }
		\STATE flag$\_$missed$\_$target = 0
		\ENDIF	
		\ENDIF	
		
	\end{algorithmic}
\end{algorithm}

\subsection{Latching to a "dummy" boat - indoors}

The tests were performed in a swimming pool, $20m \times 10m \times 1.5m$. The water was calm with minimal disturbances (robot's $roll$ and $pitch$ angles$ \leq1^\circ$). On the swimming pool the "dummy" was stationary waiting to be latched. 

The "dummy" boat is a box container integrating only a pin with bearing stud and a tag. On the other hand, the robotic boat is powered and integrates a lidar to navigate medium distances $d_m$ ($2m< d_m < 100m$ ) with an accuracy in the range of $\pm100mm$. The system also includes a camera for navigation in short distances $d_s$ ($2mm< d_s < 10m$ ) with high accuracy $\pm10mm$.

\textbf{Homing, docking and garaging steps:} 
The robotic boat is located around two meters apart from the docking station and with its camera navigates towards the "dummy" boat. With the vision based controller registers the pose and orientation of the dummy's marker.

\begin{figure}[t]
	\centering
	\includegraphics[width=0.45\textwidth]{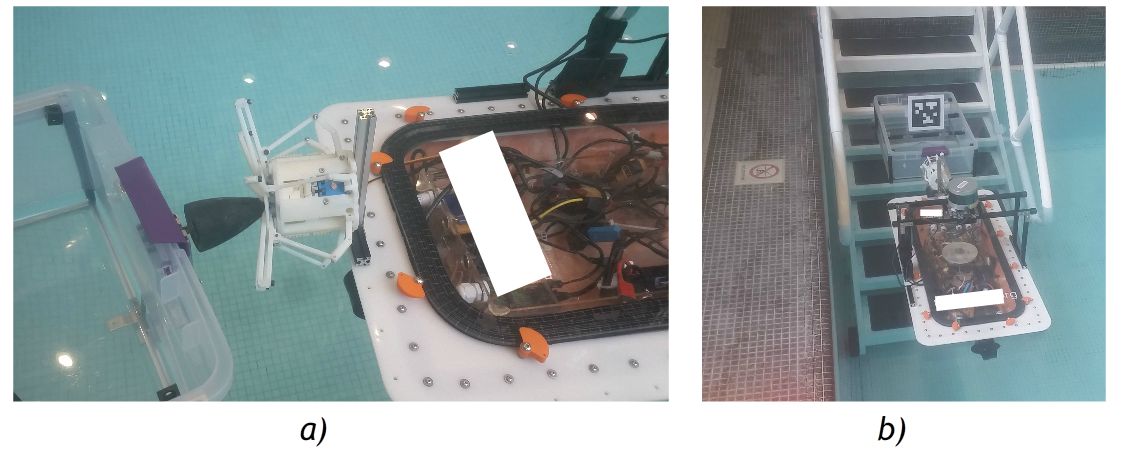}
	\includegraphics[width=0.45\textwidth]{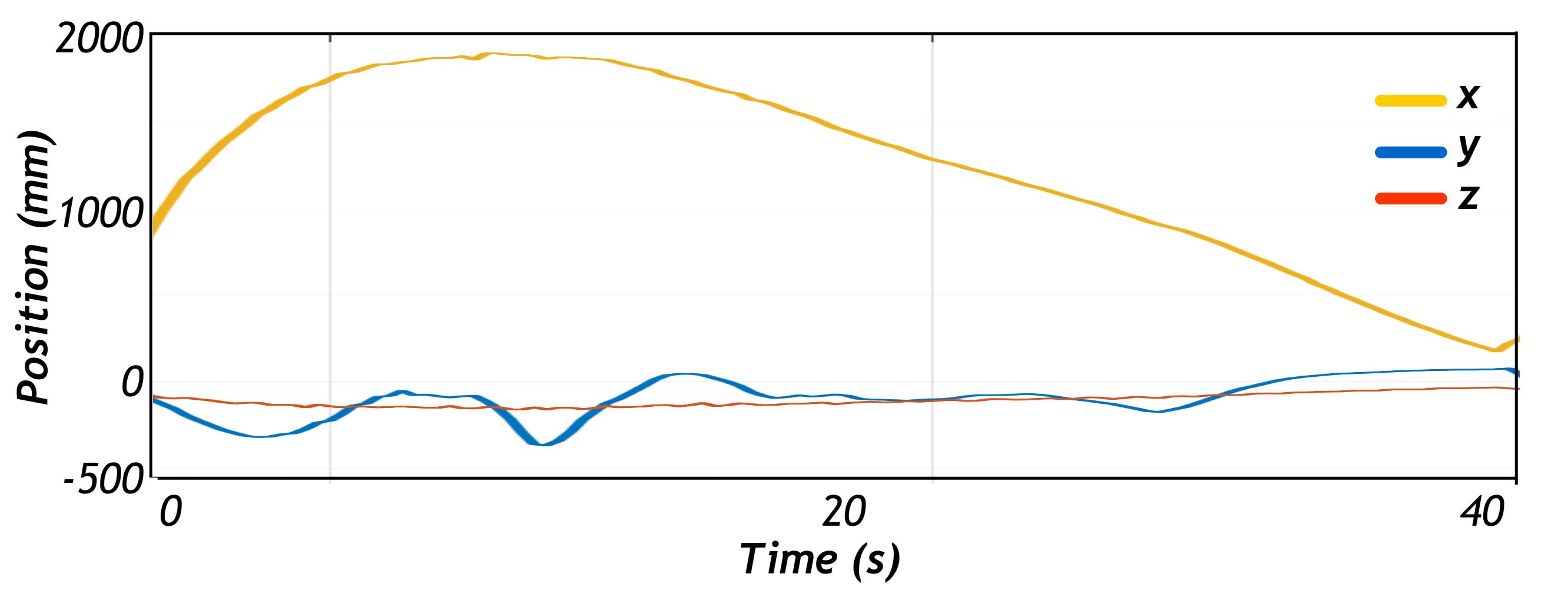}		
	\includegraphics[width=0.45\textwidth]{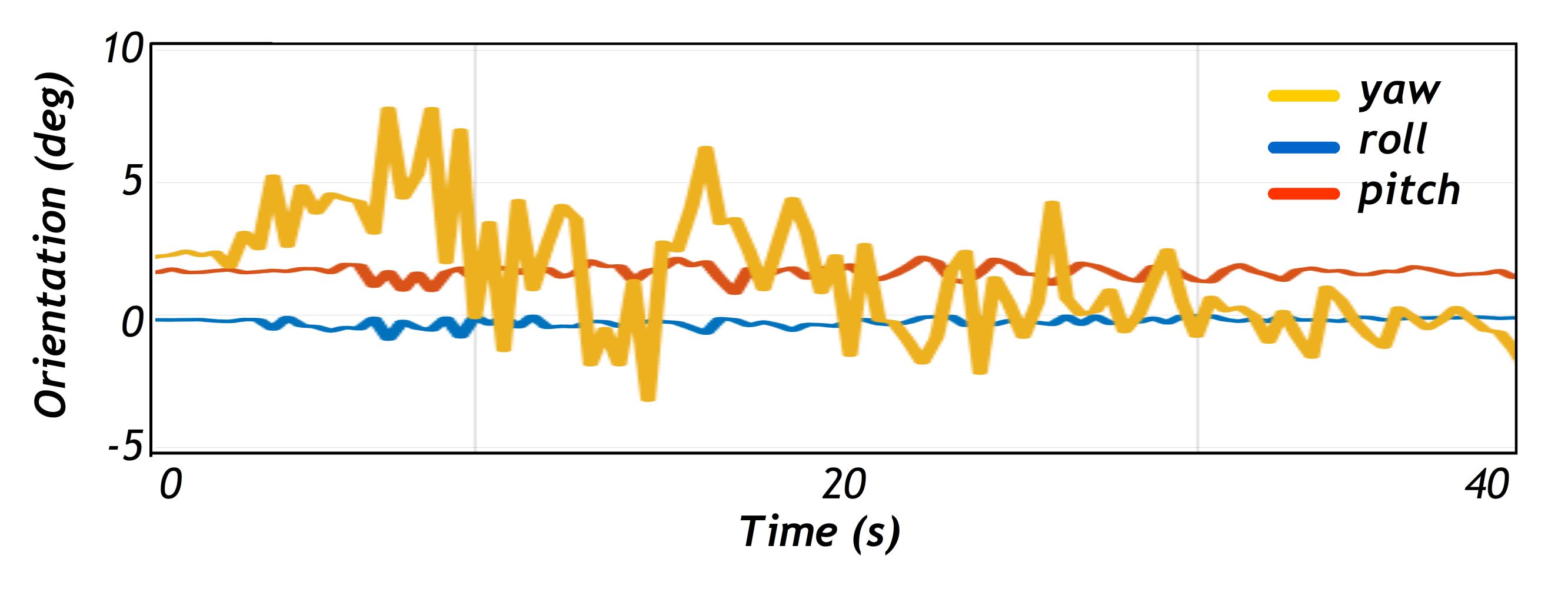}
	\includegraphics[width=0.45\textwidth]{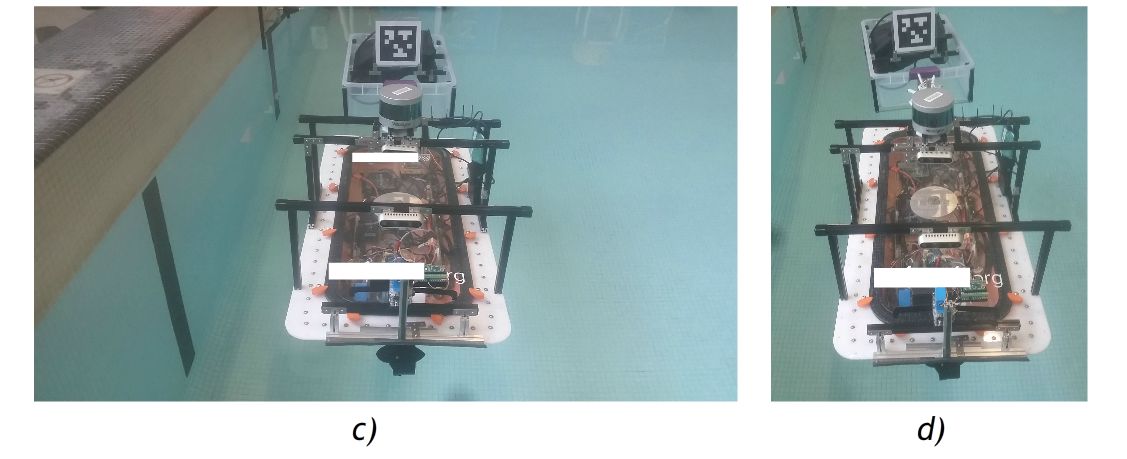}
	\includegraphics[width=0.45\textwidth]{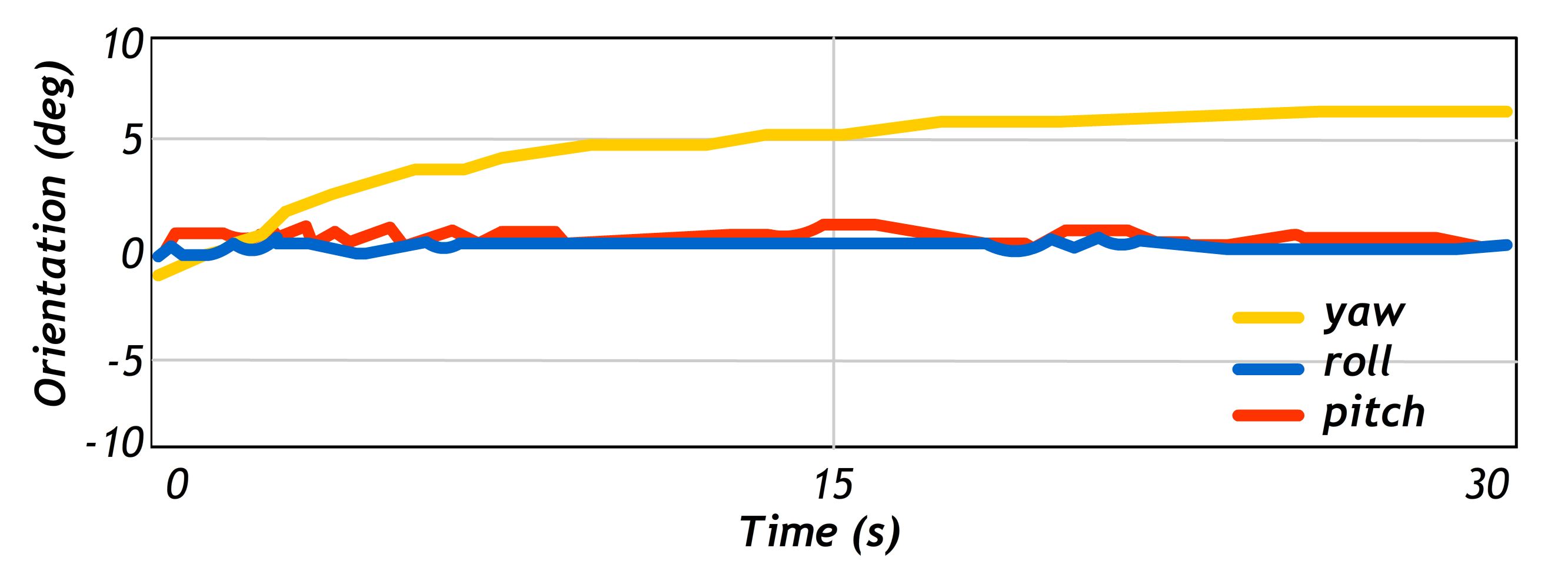}
	\caption{ a) The adaptive funnel is closing while approaching the pin on the "dummy" boat. b) The adaptive funnel latches the "dummy" closing its "tentacles". The plots show the position and orientation when latching. c) The robotic boat drives the "dummy" in straight line with the adaptive funnel fully closed. d) The 2 DoF moves the latched "dummy" $6^\circ$ to the left.}
	\label{figrp}
\end{figure}

In the joining process, the initial state of the adaptive funnel is with the "tentacles" open. While, reaching to the target's pin on the "dummy" boat, it progressively closes the "tentacles", see Figure \ref{figrp}$a$. 
Once the ball is detected inside the adaptive funnel the socket is closed, creating the spherical joint between the entities, see Figure \ref{figrp}$b$.  

\textbf{Towing in straight line:}  
The robotic boat is able to tow the "dummy" in straight line when the adaptive mechanism is set to compress the soft rubber from the target's pin, see Figure \ref{figrp}. This case scenario works for both adaptive latching systems (1 DoF or 2 Dof). 

\textbf{Towing with heading angle the power-less boat:} 
The adaptive latching with 2 Dof is able to latch and tow the power-less boat with a certain heading angle. In the experiment the robotic boat is able to tow the "dummy" while keeping an angle $6^\circ$ between them, see Figure \ref{figrp}$d$.


\begin{figure}[t]
	\centering
	\includegraphics[width=0.49\textwidth]{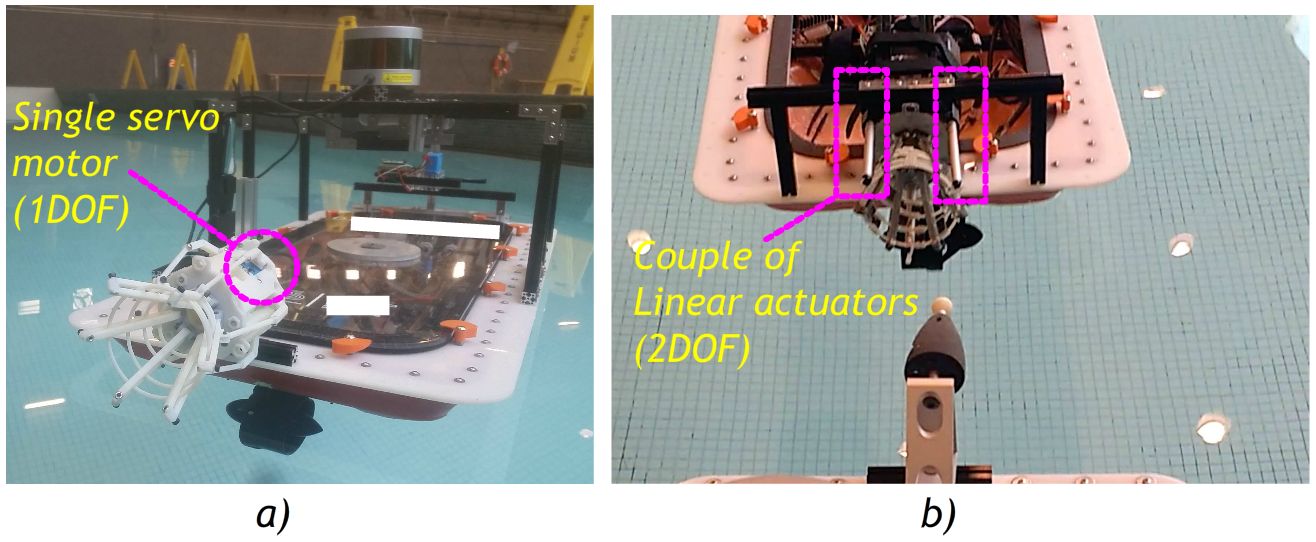}
	\caption{ a) 1 DoF adaptive funnel with one actuator moving all the arms symmetrically. b) 2 DoF adaptive latching with a couple of linear actuators opposite to each other enables asymmetrical  configurations for guiding.} 
	\label{dof}
\end{figure}

\subsection{Latching to a robotic boat - outdoors}

On the river, we tested the most challenging latching, which is between two floating entities. We tested with the 1 DoF latching system. The water and wind were calm making the robot's $pitch$ and $roll$ angles up to $\pm 1.5^\circ$. 

In the experiment, the balancing robotic boat (the target robot $T$) balances computing a NDT matching from the lidar sensor. The registered error in position was in the range of $\pm 100mm$ and the error in orientation: $roll$  $\pm 1^\circ$, $pitch$ $\pm 1.5^\circ$ and a $yaw$ $\pm 2^\circ$. 

\begin{figure}[t]
	\centering
	\includegraphics[width=0.45\textwidth]{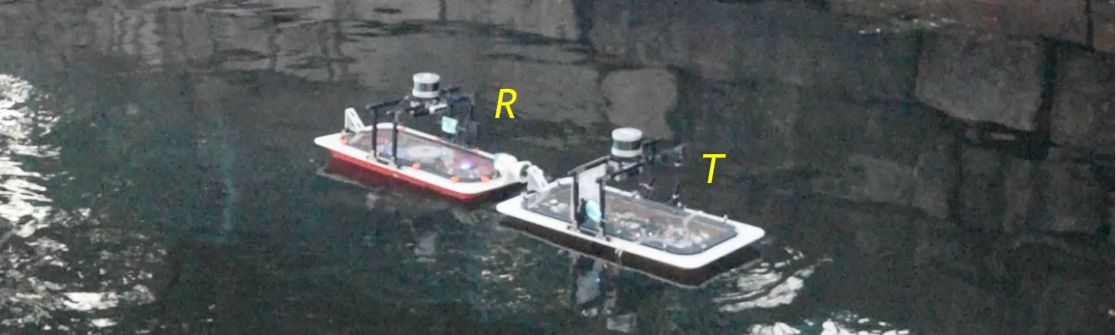}
	\includegraphics[width=0.45\textwidth]{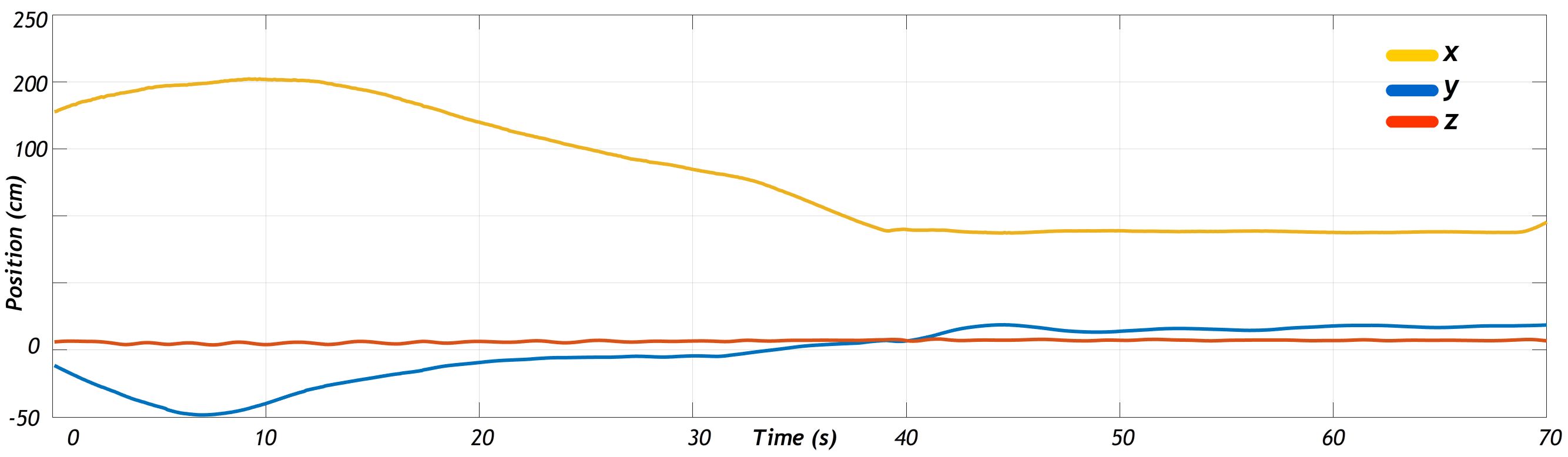}
	\includegraphics[width=0.45\textwidth]{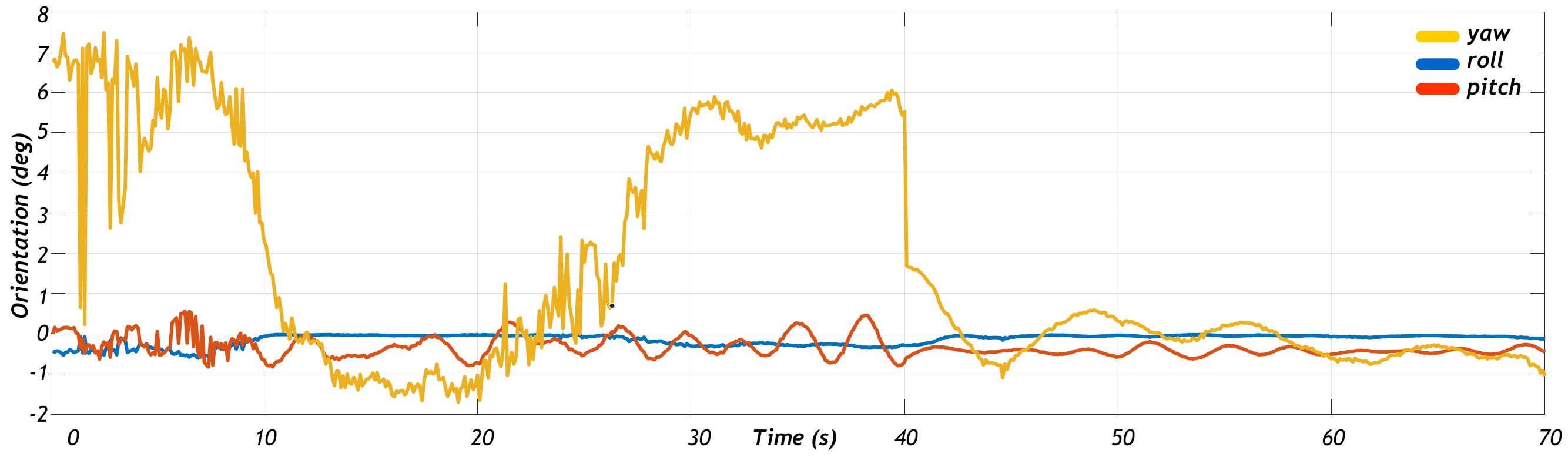}
	
	\caption{Adaptive funnel with 1 DoF for the use case of latching and towing (top). The robotic boat $R$ starts tracking the target $T$ from  around two meters apart, the entities are latched when $x=89cm$, $d_y < \pm 4cm$ and $yaw < \pm 27.5^\circ$ (middle) and orientation $RPY\leq1^\circ$ (bottom). 
	The latching process is performed in time 1 to 40, and the towing is from 41 to 70 seconds.}
	\label{figpherical3}
\end{figure}

Figure \ref{figpherical3} shows the experimental results from the  test on the river. The  robotic boat $R$ with adaptive funnel wants to latch to the balancing robot with target $T$. The plots reveal the position and orientation for latching and towing.  
In these experiments, the robotic boats are latched when the position in $d_x < 900mm$, $d_y < \pm 40mm$ and $yaw < \pm 27.5^\circ$. Once the robots are latched, robot $R$ tows in straight line robot $T$. The plot shows the waves in the $pitch$ readings before and after the latching. When the latching is secured and  robot $R$ starts to tow robot $T$, it is possible to notice an attenuation in $RPY\leq\pm1^\circ$. 

\section{Conclusion and future work}
\label{sec:conclusion}

The paper presents a novel bio-inspired adaptive latching system for towing and guiding floating platforms. The system is based on the squid animal, which is able to trap and apply multiple forces to adjust the position of the holding object with its tentacles.

The adaptive latching system for robotic boats is able to latch efficiently to a dock, to another robotic boat or to a floating structure. The system is mechanically reliable, based on the spherical joint principle that enables rotation and movement in two direction between the latched parts. 

The system integrates a visual controller based on the camera-tag principle to position the floating entities ready to latch. Further, the vision based controller is able to guide a latched power-less "dummy" boat with no sensors and no thrusters. This is useful in  real case scenarios in the canals of Amsterdam, for directing the "dummy" boat in narrow spaces (narrow canals) preventing it to hit the walls and for guiding it when turning in a curve.

The experiments were performed in indoors swimming pool as well as outdoors, on the Charles river and Amsterdam canals. 
The robotic boats are able to connect while overcoming water disturbances and misalignment. Also, the connection is strong to tow and guide a power-less "dummy" boat or to tow a similar robot. 


\section*{Acknowledgment}
We would like to thank the grant from the Amsterdam Institute for Advanced Metropolitan Solutions (AMS) in Netherlands to support this research. 

\addtolength{\textheight}{-3cm}   



\bibliographystyle{IEEEtran}
\bibliography{af}

\end{document}